\documentclass{article}
\usepackage{blindtext}
\usepackage{graphicx}
\usepackage{array}
\usepackage[numbers,sort&compress]{natbib}
\usepackage{url}
\usepackage{multirow}
\usepackage{spconf,amsmath,epsfig,subfig}
\usepackage{hyperref} 
\usepackage{comment}

\usepackage[table]{xcolor}
\usepackage{adjustbox}
\usepackage[geometry]{ifsym}
\usepackage{enumitem}
\usepackage{siunitx}
\usepackage{bm}
\usepackage{wrapfig}

\usepackage{color,soul}

\title{Object Recognition Under Multifarious Conditions: A Reliability Analysis and A Feature Similarity-Based Performance Estimation}

\name{Dogancan Temel*\thanks{*Equal contribution.}, Jinsol Lee*\thanks{Dataset: \url{https://ghassanalregib.com/cure-or/}}, and Ghassan AlRegib}
\address{Center for Signal and Information Processing,\\ School of Electrical and Computer Engineering,\\ Georgia Institute of Technology, Atlanta, GA, 30332-0250\\ \{cantemel, jinsol.lee, alregib\}@gatech.edu}

\begin{document}

\onecolumn 

\begin{description}[labelindent=1cm,leftmargin=3cm,style=multiline]

\item[\textbf{Citation}]{D. Temel, J. Lee and G. AlRegib, "Object Recognition Under Multifarious Conditions: A Reliability Analysis and A Feature Similarity-Based Performance Estimation," IEEE International Conference on Image Processing (ICIP), Taipei, Taiwan, 2019.
} \\



\item[\textbf{Dataset}]{\url{https://ghassanalregib.com/cure-or/}} \\

\item[\textbf{Bib}] {
@INPROCEEDINGS\{Temel2019\_ICIP,\\ 
author=\{D. Temel and J. Lee and G. AIRegib\},\\ 
booktitle=\{IEEE International Conference on Image Processing (ICIP)\},\\ 
title=\{Object Recognition Under Multifarious Conditions: A Reliability Analysis and A Feature Similarity-Based Performance Estimation\},\\ 
year=\{2019\},\}\\
} \\

\item[\textbf{Copyright}]{\textcopyright 2019 IEEE. Personal use of this material is permitted. Permission from IEEE must be obtained for all other uses, in any current or future media, including reprinting/republishing this material for advertising or promotional purposes,
creating new collective works, for resale or redistribution to servers or lists, or reuse of any copyrighted component
of this work in other works. } \\

\item[\textbf{Contact}]{\href{mailto:alregib@gatech.edu}{alregib@gatech.edu}~~~~~~~\url{https://ghassanalregib.com/}\\ \href{mailto:dcantemel@gmail.com}{dcantemel@gmail.com}~~~~~~~\url{http://cantemel.com/}}
\end{description} 

\thispagestyle{empty}
\newpage
\clearpage

\twocolumn

\maketitle


\begin{abstract} 
In this paper, we investigate the reliability of online recognition platforms, Amazon Rekognition and Microsoft Azure, with respect to changes in background, acquisition device, and object orientation. We focus on platforms that are commonly used by the public to better understand their real-world performances. To assess the variation in recognition performance, we perform a controlled experiment by changing the acquisition conditions one at a time. We use three smartphones, one DSLR, and one webcam to capture side views and overhead views of objects in a living room, an office, and photo studio setups. Moreover, we introduce a framework to estimate the recognition performance with respect to backgrounds and orientations. In this framework, we utilize both handcrafted features based on color, texture, and shape characteristics and data-driven features obtained from deep neural networks. Experimental results show that deep learning-based image representations can estimate the recognition performance variation with a Spearman's rank-order correlation of 0.94 under multifarious acquisition conditions.


\end{abstract}

\vspace{-1mm}
\begin{keywords}
object dataset, controlled experiment with recognition platforms, performance estimation, deep learning, feature similarity
\end{keywords}
\vspace{-2mm}

\maketitle

\vspace{-2mm}

\section{Introduction}
\begin{figure}[h]
\begin{minipage}[b]{0.19\linewidth}
  \centering
\includegraphics[width=\textwidth]{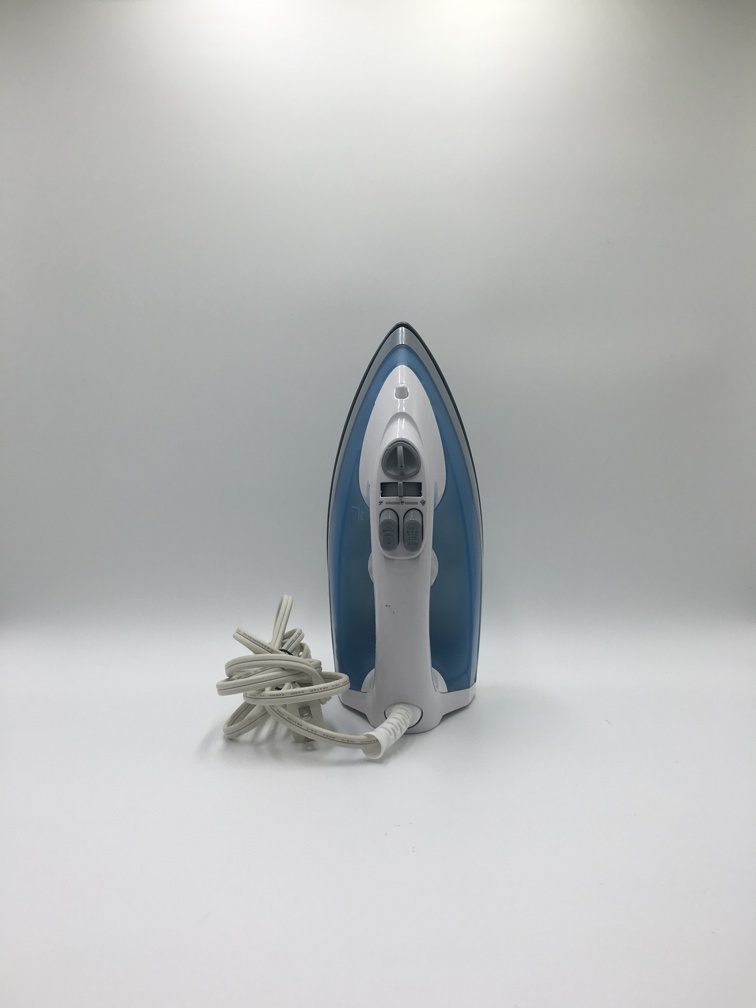}
  \centerline{\footnotesize{(a) {\tabular[t]{@{}l@{}}White\\  \endtabular}}}
\end{minipage}
\vspace{.2mm}
\hfill
\begin{minipage}[b]{0.19\linewidth}
  \centering
\includegraphics[width=\textwidth]{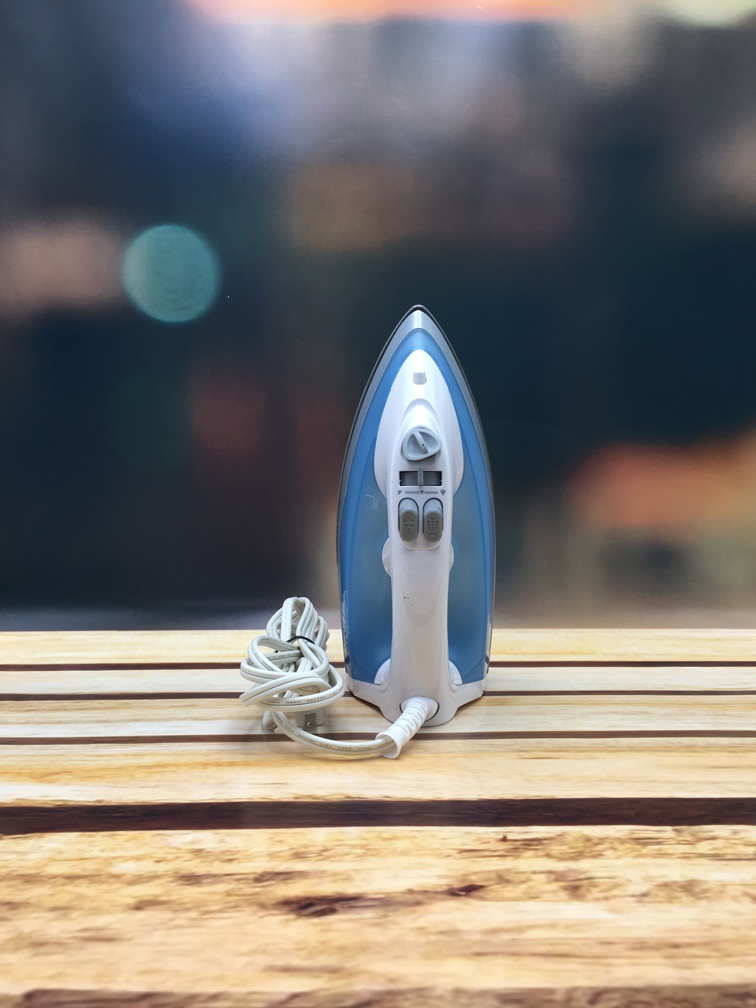}
  \centerline{\footnotesize{(b) {\tabular[t]{@{}l@{}}2D Living\\Room \endtabular}}}
\end{minipage}
\vspace{.2mm}
\hfill
\begin{minipage}[b]{0.19\linewidth}
  \centering
\includegraphics[width=\textwidth]{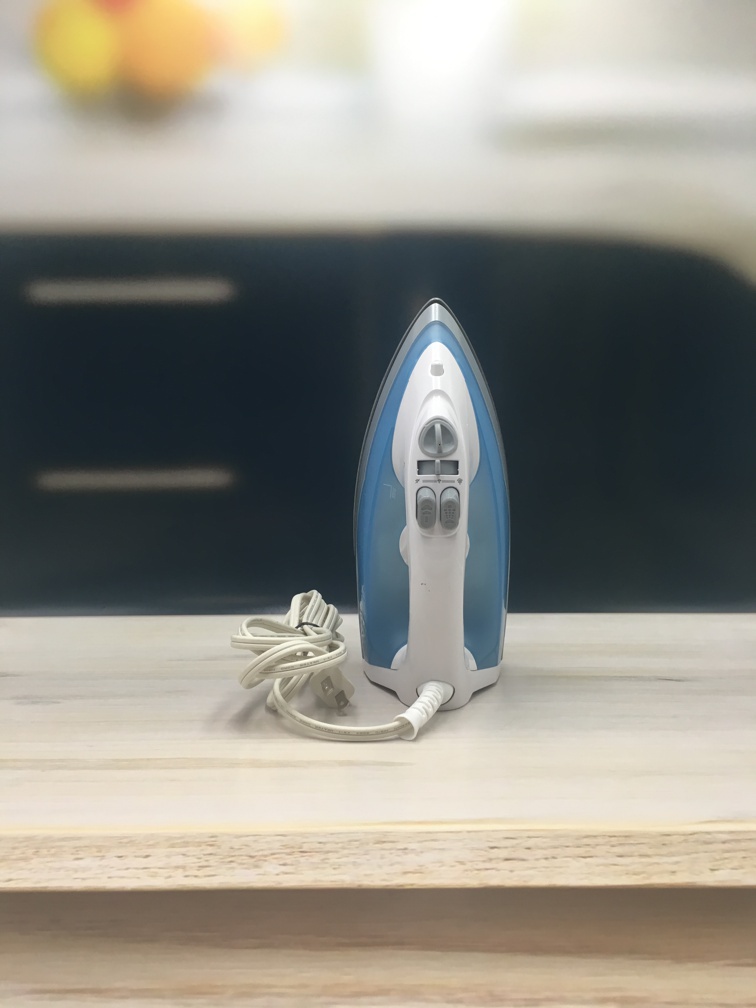}
  \centerline{\footnotesize{(c) {\tabular[t]{@{}l@{}}2D Kitchen\\  \endtabular}}}
\end{minipage}
\vspace{.2mm}
\hfill
\begin{minipage}[b]{0.19\linewidth}
  \centering
\includegraphics[width=\textwidth]{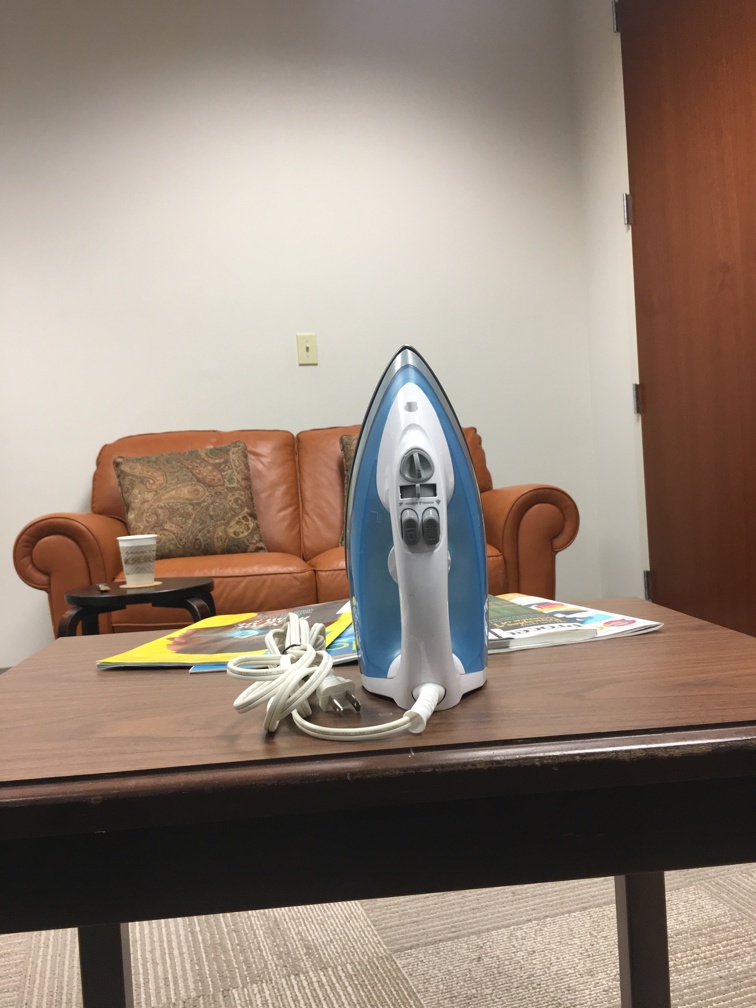}
  \centerline{\footnotesize{(d) {\tabular[t]{@{}l@{}}3D Living\\Room  \endtabular}}}
\end{minipage}
\vspace{.2mm}
\hfill
\begin{minipage}[b]{0.19\linewidth}
  \centering
\includegraphics[width=\textwidth]{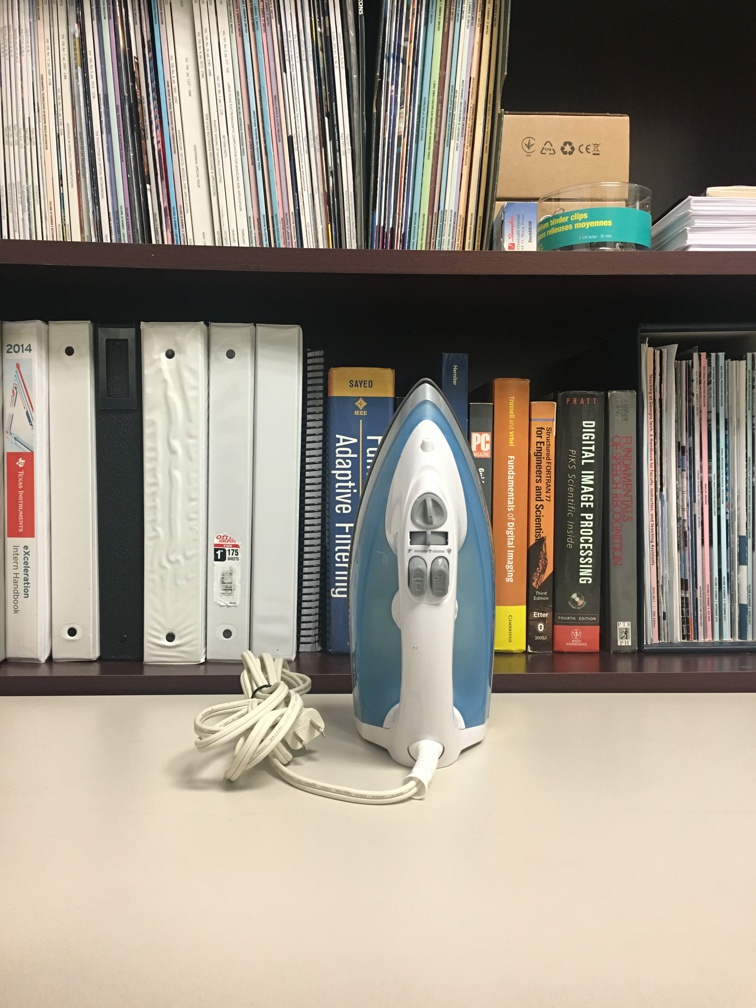}
  \centerline{\footnotesize{(e) {\tabular[t]{@{}l@{}}3D Office\\  \endtabular}}}
\end{minipage}
\vspace{.2mm}
\hfill
\begin{minipage}[b]{0.19\linewidth}
  \centering
\includegraphics[width=\textwidth]{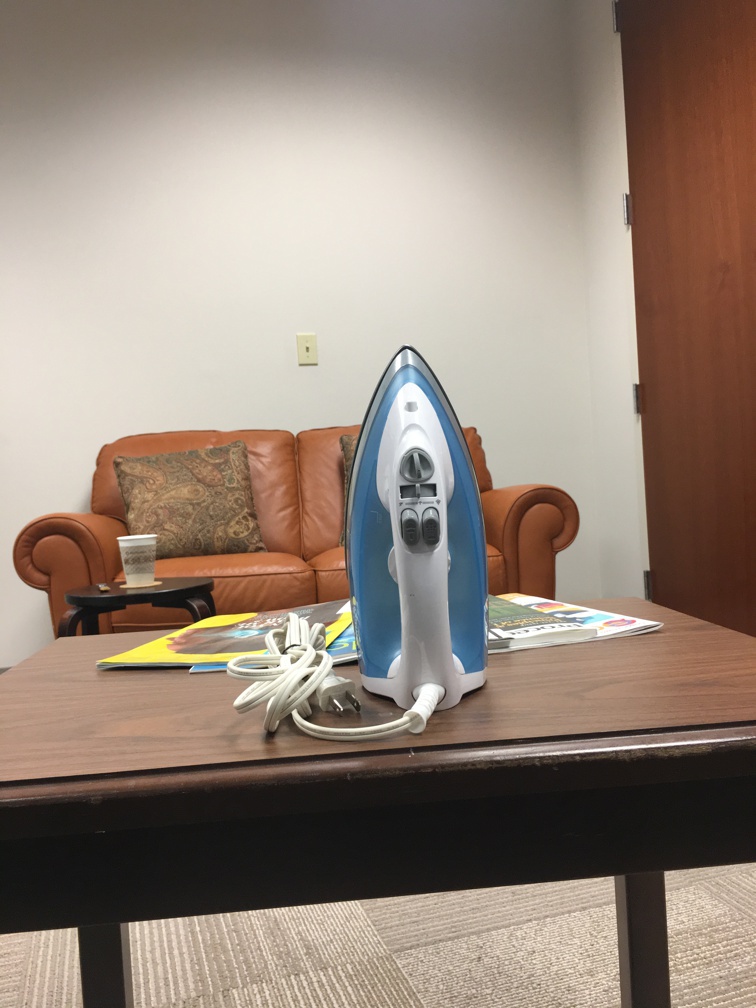}
  \centerline{\footnotesize{(f) \ang{0} (Front) }}
\end{minipage}
\hfill
\begin{minipage}[b]{0.19\linewidth}
  \centering
\includegraphics[width=\textwidth]{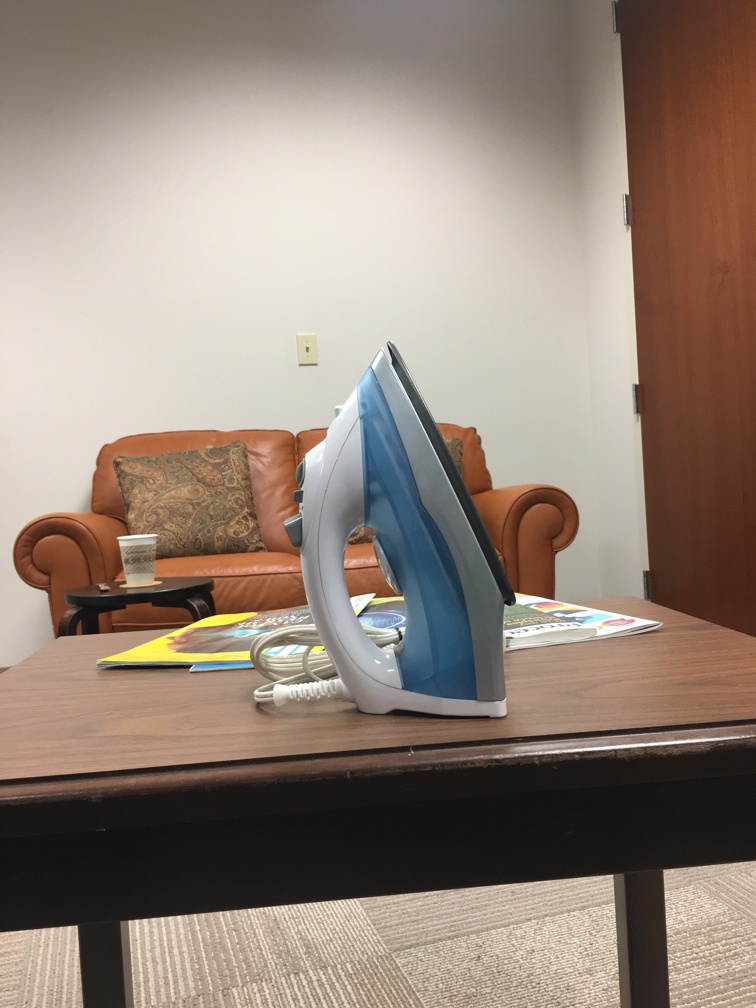}
  \centerline{\footnotesize{(g) \ang{90}}}
\end{minipage}
\hfill
\begin{minipage}[b]{0.19\linewidth}
  \centering
\includegraphics[width=\textwidth]{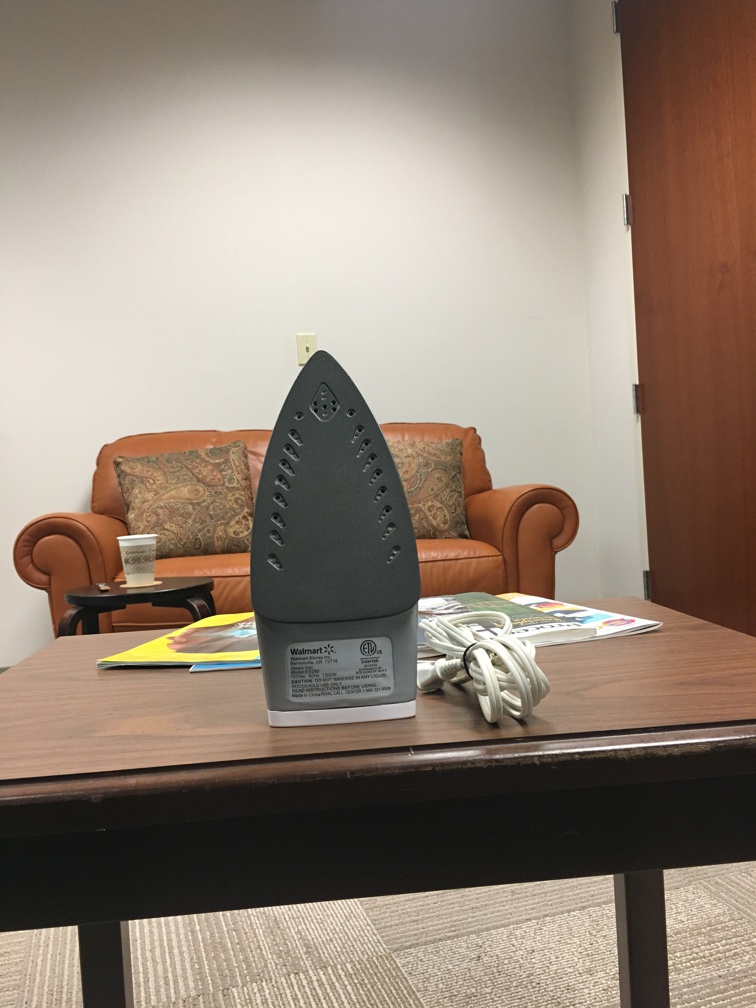}
  \centerline{\footnotesize{(h) \ang{180}}}
\end{minipage}
\hfill
\begin{minipage}[b]{0.19\linewidth}
  \centering
\includegraphics[width=\textwidth]{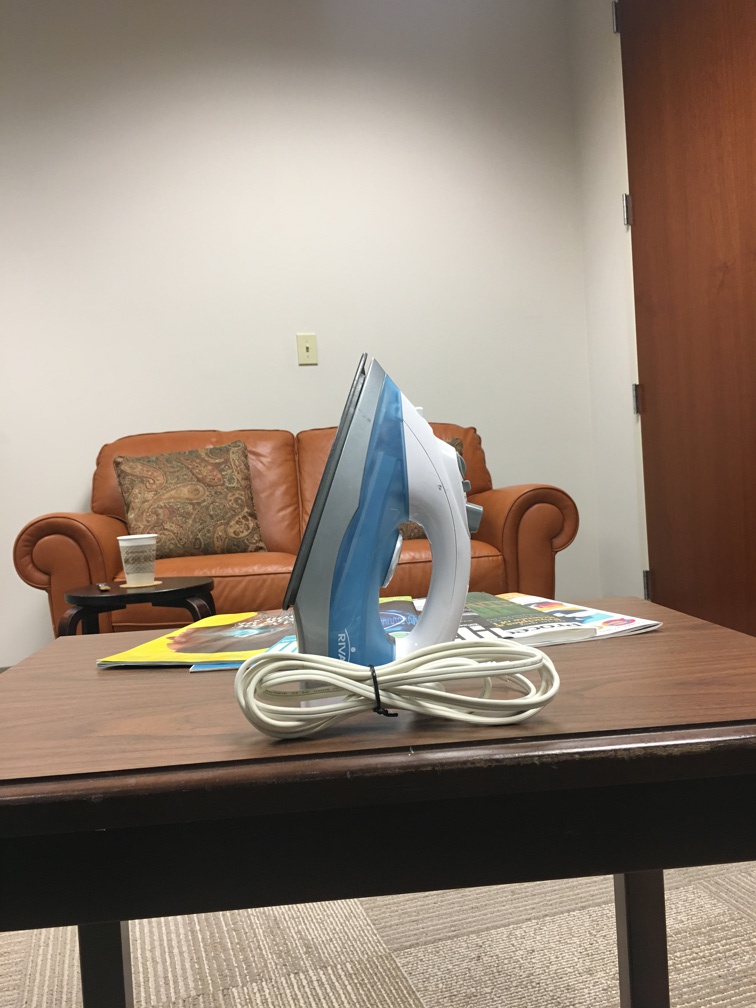}
  \centerline{\footnotesize{(i) \ang{270}}}
\end{minipage}
\hfill
\begin{minipage}[b]{0.19\linewidth}
  \centering
\includegraphics[ width=\textwidth]{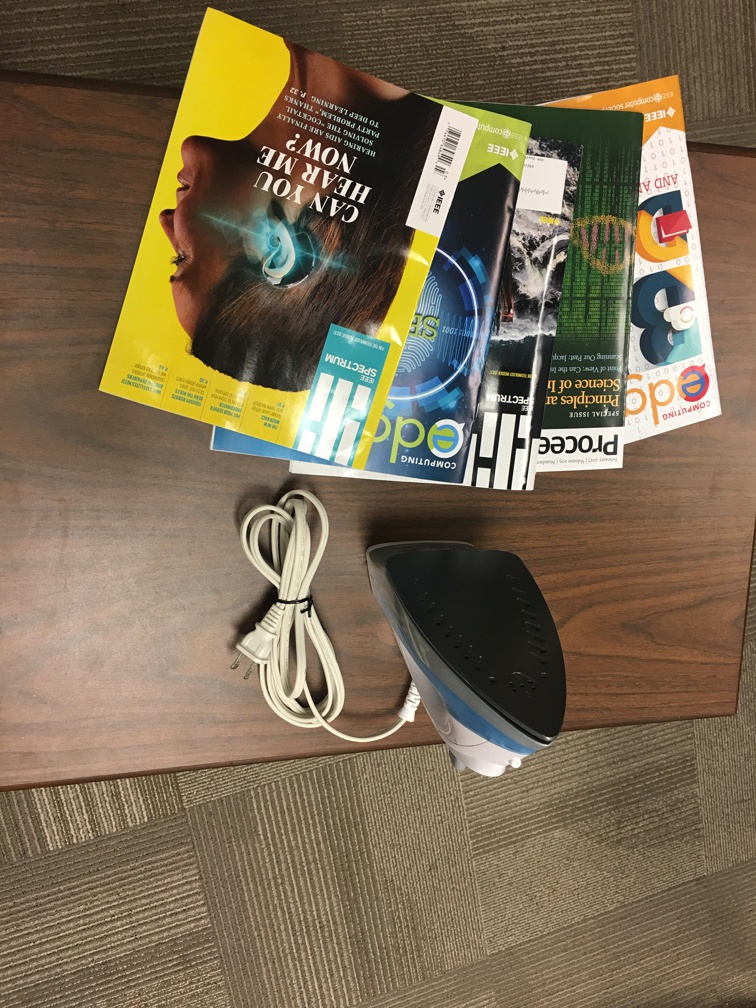}
  \centerline{\footnotesize{(j) Overhead}}
\end{minipage}
\vspace{-1mm}
\caption{Object backgrounds and orientations in \texttt{CURE-OR}.}
\vspace{-2mm}
\label{fig:images}
\end{figure}

\vspace{-2mm}
In recent years, the performance of visual recognition and detection algorithms have considerably advanced with the progression of data-driven approaches and computational capabilities \cite{Deng2009, Lin2014}. These advancements enabled state-of-the-art methods to achieve human-level performance in specific recognition tasks \cite{He2015,Wu2015}. Despite these significant achievements, it remains a challenge to utilize such technologies in real-world environments that diverge from training conditions. To identify the factors that can affect recognition performance, we need to perform controlled experiments as in  \cite{Dodge2016,Zhou2017,Hosseini2017,Lu2017,Das2018}. Even though these studies shed a light on the vulnerability of existing recognition approaches, investigated conditions are either limited or unrealistic. Recently, we introduced the \texttt{CURE-OR} dataset and analyzed the recognition performance with respect to simulated challenging conditions \cite{Temel2018_CUREOR}. Hendrycks and Dietterich \cite{Hendrycks2019} also studied the effect of similar conditions by postprocessing the images in ImageNet \cite{Deng2009}. In \cite{Temel2017_NIPSW,Temel2018_SPM,Prabhushankar2018_SemanticFilters,Temel2019_VIP}, performance variation under simulated challenging conditions were analyzed for traffic sign recognition and detection. Aforementioned studies overlooked the acquisition conditions and investigated the effect of simulated conditions. \emph{In contrast to the literature \cite{Dodge2016,Zhou2017,Hosseini2017,Lu2017,Das2018,Temel2017_NIPSW,Temel2018_SPM,Temel2019_VIP,Hendrycks2019} and our previous work \cite{Temel2018_CUREOR}, the main focus of this study is to analyze the effect of real-world acquisition conditions including device type, orientation and background.} In Fig.~\ref{fig:images}, we show sample images obtained under different acquisition conditions.

If we consider ideal acquisition conditions as reference conditions that lead to the highest recognition rate, any variation would decrease the recognition performance and affect visual representations. \emph{Based on this assumption, we hypothesize that recognition performance variations can be estimated by variations in visual representations.} Overall, the contributions of this manuscript are five folds. First, we investigate the effect of \emph{background} on object recognition by performing controlled experiments with different backgrounds. Second, we analyze the effect of \emph{acquisition devices} by comparing the recognition accuracy of images captured with different devices. Third, we analyze the recognition performance with respect to different \emph{orientation} configurations. Fourth, we introduce a framework to \emph{estimate} the recognition \emph{performance} variation under varying backgrounds and orientations. Fifth, we benchmark the performance of \emph{handcrafted} and \emph{data-driven} features obtained from \emph{deep neural networks} in the proposed framework. The outline of this paper is as follows. In Section~\ref{sec:main_recognition}, we analyze the objective recognition performance with respect to acquisition conditions. In Section~\ref{sec:main_estimation}, we describe the recognition performance estimation framework and benchmark hand-crafted and data-driven methods. Finally, we conclude our work in Section~\ref{sec:conc}.



\label{sec:intro}

\vspace{-2mm}
\section{Recognition Under Multifarious Conditions}
\vspace{-2mm}
\label{sec:main_recognition}
Based on scalability, user-friendliness, computation time, service fees, access to labels and confidence scores, we assessed off-the-shelf platforms and decided to utilize Microsoft Azure Computer Vision (MSFT) and Amazon Rekognition (AMZN) platforms. As a test set, we use the recently introduced \texttt{CURE-OR} dataset that includes one million images of 100 objects captured with different devices under various object orientations, backgrounds, and simulated challenging conditions. Objects are classified into 6 categories: toys, personnel belongings, office supplies, household items, sport/entertainment items, and health/personal care items as described in \cite{Temel2018_CUREOR}. We identified 4 objects per category for each platform for testing, but because Azure only identified 3 objects correctly in one category, we excluded an object with the lowest number of correctly identified images from Amazon for fair comparison. Therefore, we used $23$ objects while assessing the robustness of the recognition platforms. Original images (challenge-free) in each category were processed to simulate realistic challenging conditions including underexposure, overexposure, blur, contrast, dirty lens, salt and pepper noise, and resizing as illustrated in \cite{Temel2018_CUREOR}. We calculated the top-5 accuracy for each challenge category to quantify recognition performance. Specifically, we calculated the ratio of correct classifications for each object in which ground truth label was among the highest five predictions.

\begin{figure}[htbp!]
\vspace{-9mm}
\begin{minipage}[b]{0.46\linewidth}
  \centering
\includegraphics[width=\textwidth]{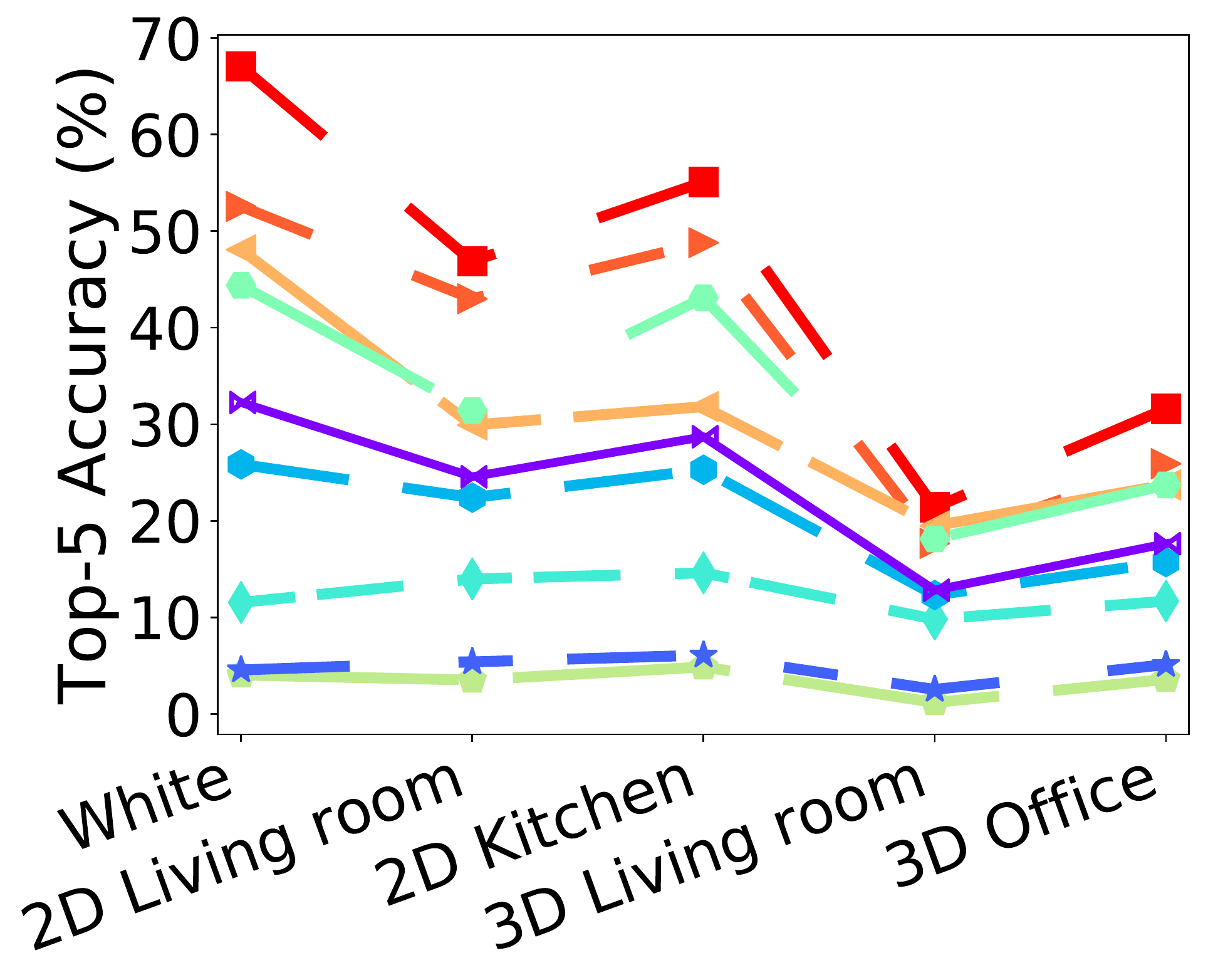}
  \vspace{0.1cm}
  \centerline{\footnotesize{(a) AMZN: Backgrounds}}
  \vspace{0.01 cm}
\end{minipage}
\vspace{0.1cm}
\begin{minipage}[b]{0.46\linewidth}
  \centering
\includegraphics[width=\linewidth]{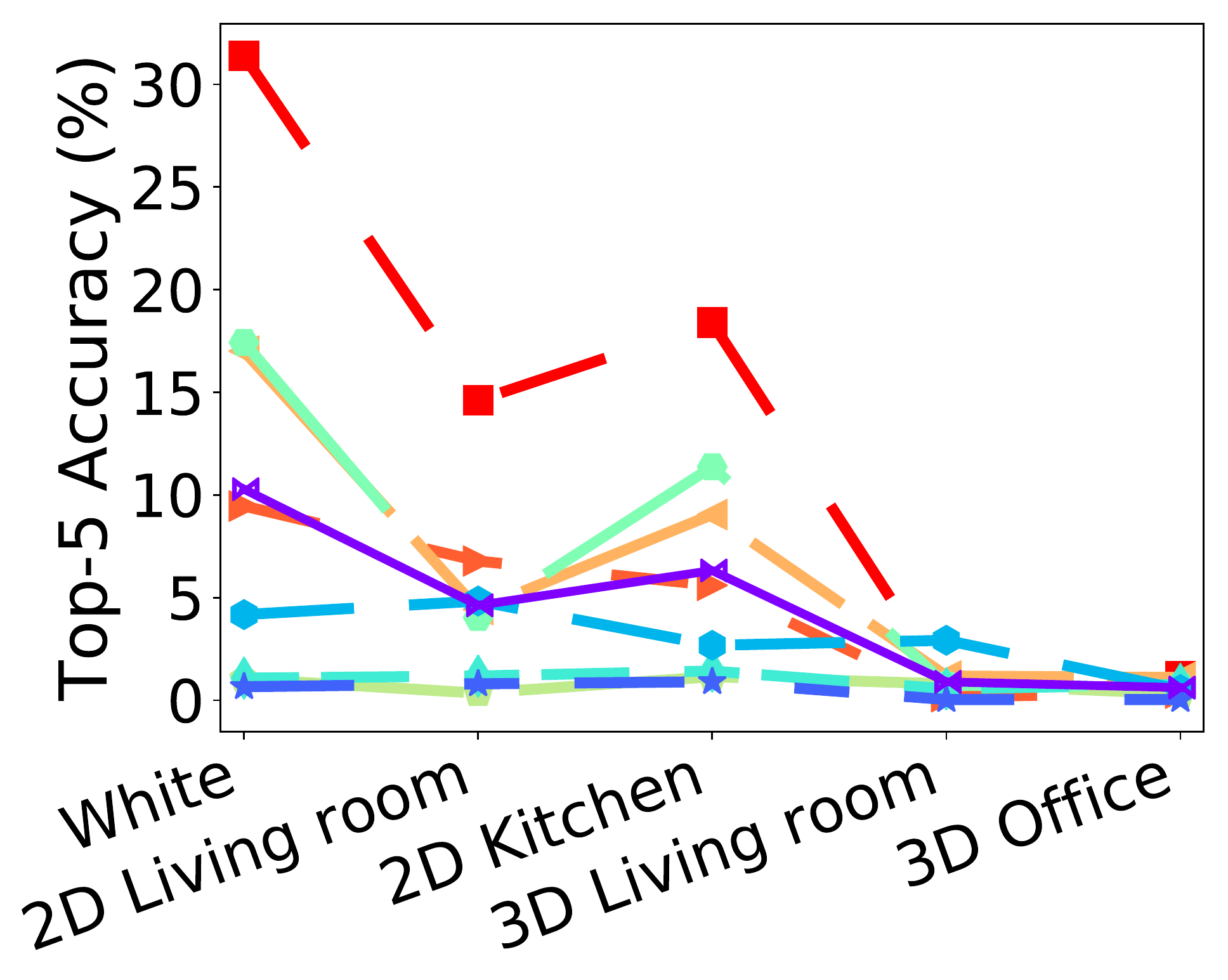}
  \vspace{0.1 cm}
  \centerline{\footnotesize{(b) MSFT: Backgrounds} }
  \vspace{0.01 cm}
\end{minipage}
\centering

\begin{minipage}[b]{0.46\linewidth}
  \centering
\includegraphics[width=\linewidth]{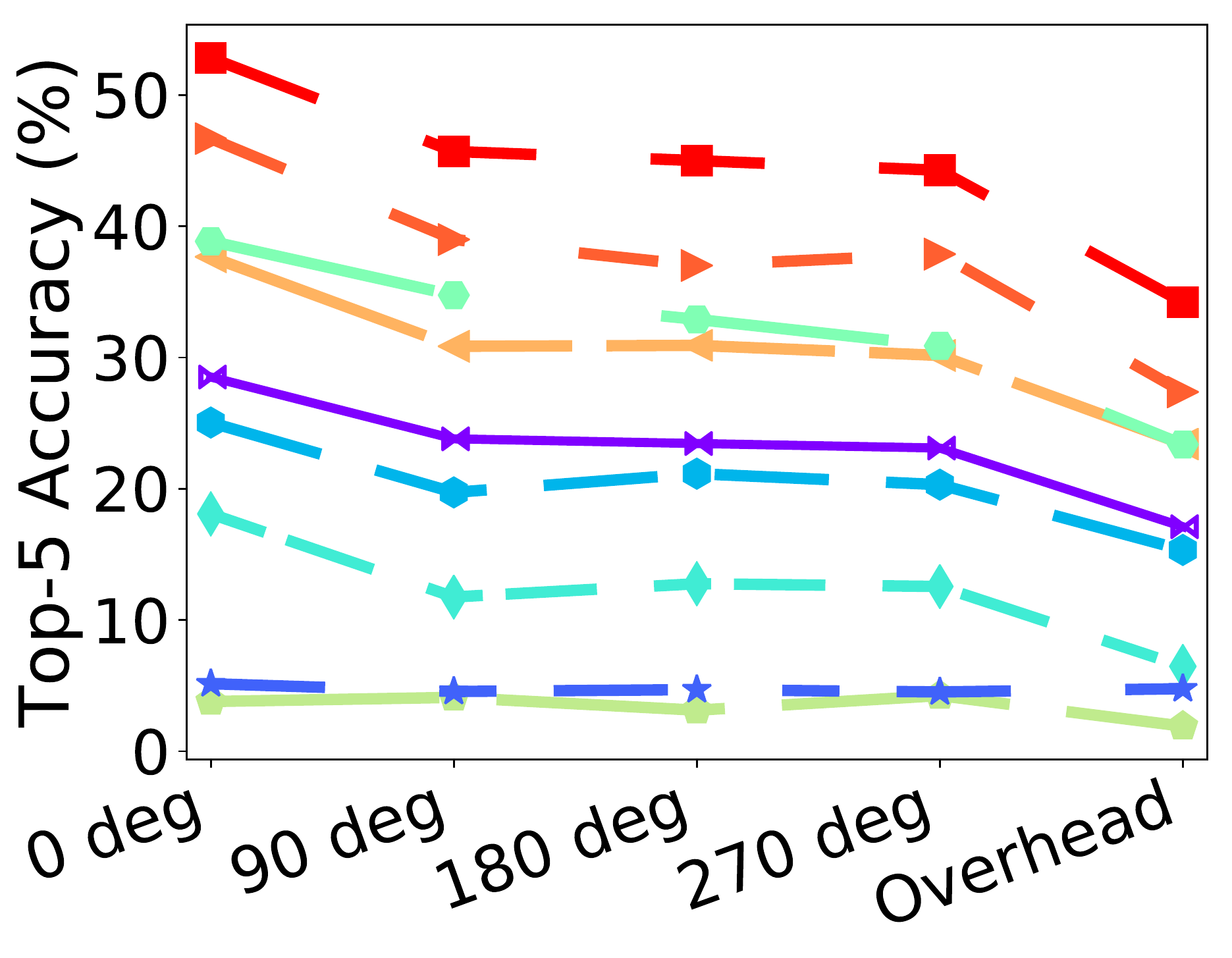}
  \vspace{0.01 cm}
  \centerline{\footnotesize{(c) AMZN: Orientations}}
  \vspace{0.01 cm}
\end{minipage}
\vspace{0.1cm}
\begin{minipage}[b]{0.46\linewidth}
  \centering
\includegraphics[width=\linewidth, trim=5mm 0mm 5mm 10mm]{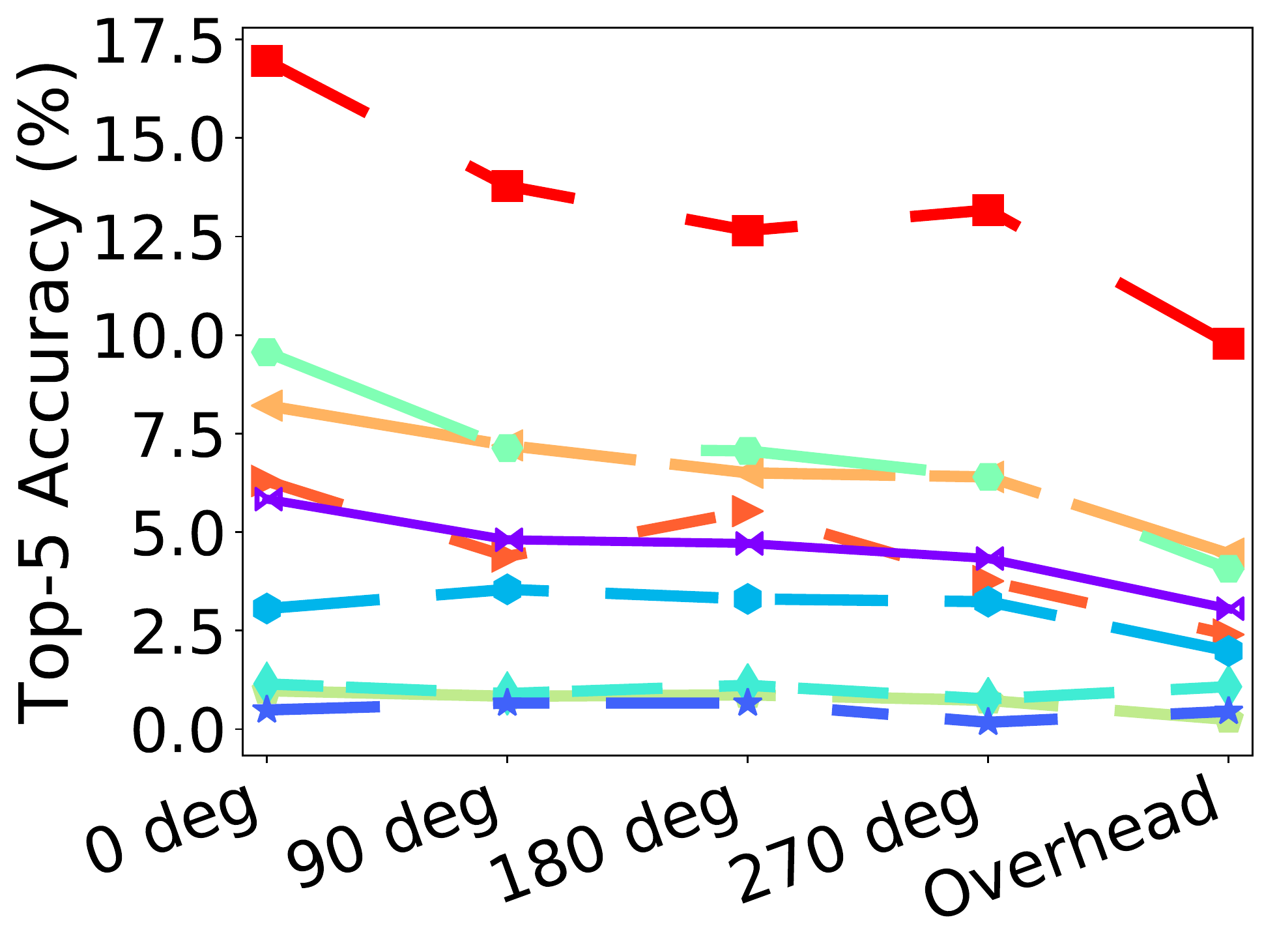}
  \vspace{0.01 cm}
  \centerline{\footnotesize{(d) MSFT: Orientations}}
  \vspace{0.01 cm}
\end{minipage}

\begin{minipage}[b]{0.46\linewidth}
  \centering
\includegraphics[width=\textwidth]{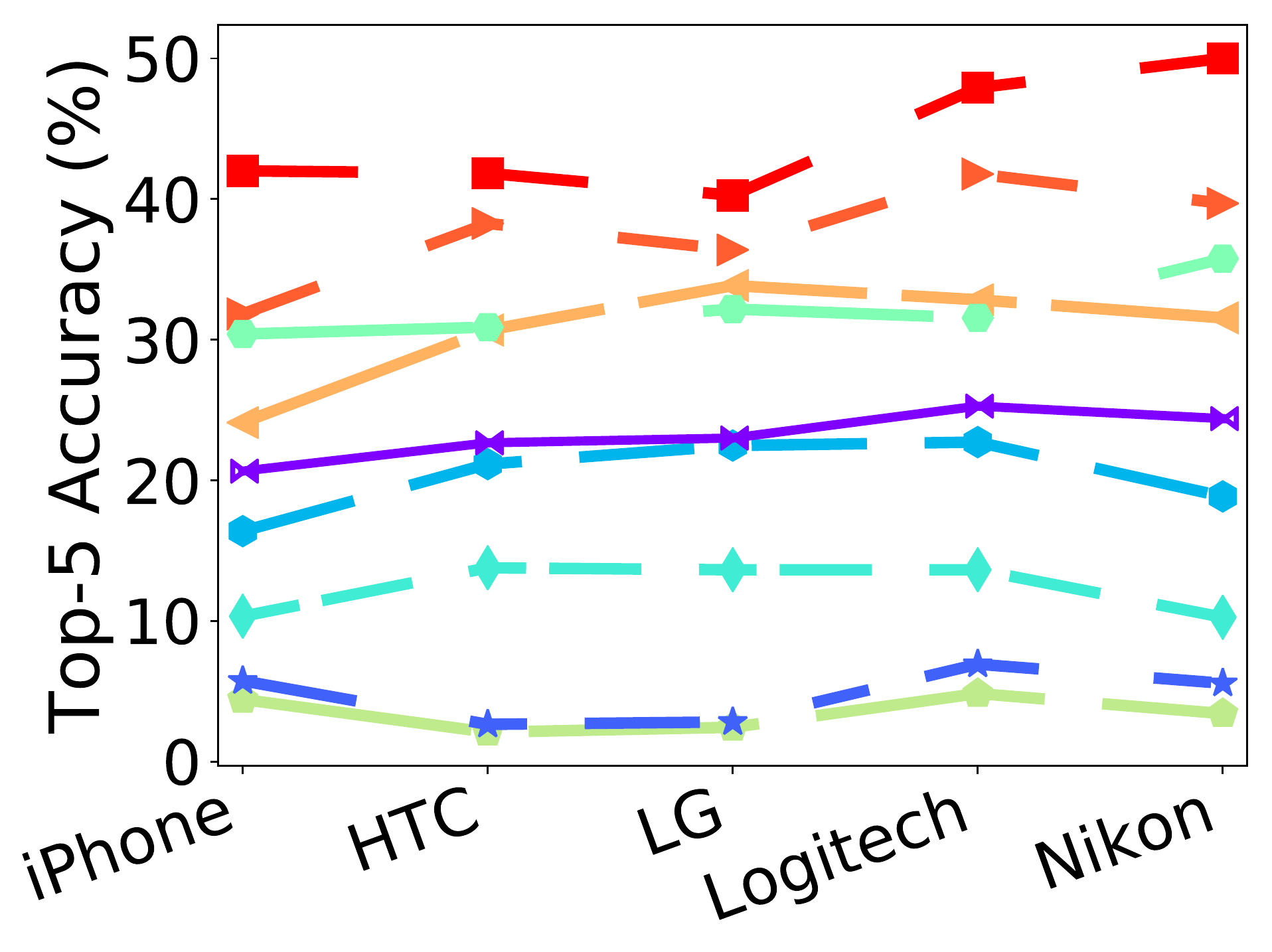}
  \vspace{1mm}
  \centerline{\footnotesize{(e) AMZN: Devices}}
\end{minipage}
\begin{minipage}[b]{0.46\linewidth}
  \centering
\includegraphics[width=\linewidth]{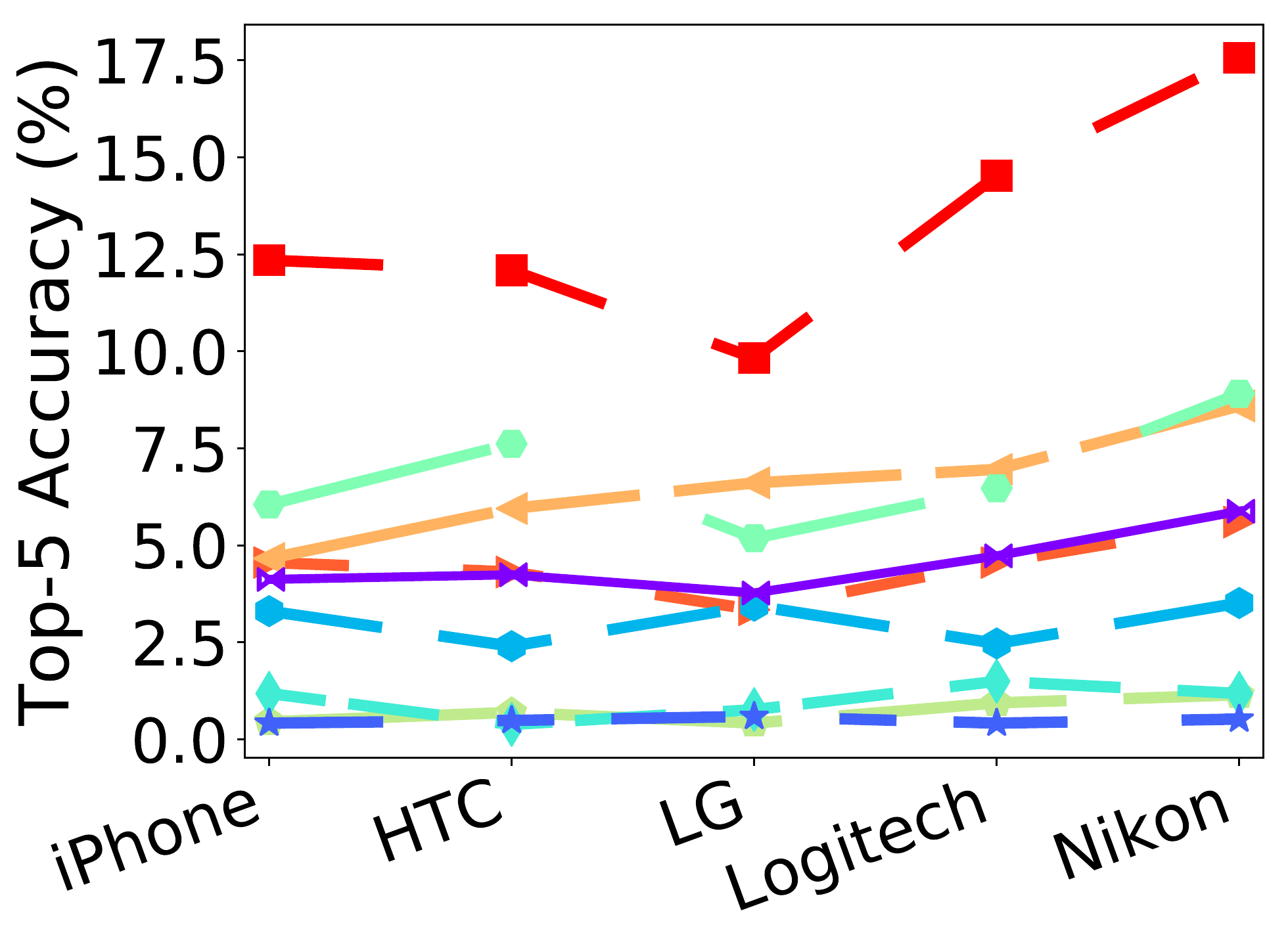}
  \vspace{1mm}
  \centerline{\footnotesize{(f) MSFT: Devices}}
\end{minipage}
\centering
  \vspace{0.5mm}

\includegraphics[width=0.8\linewidth, trim = 30mm 5mm 30mm 0mm]{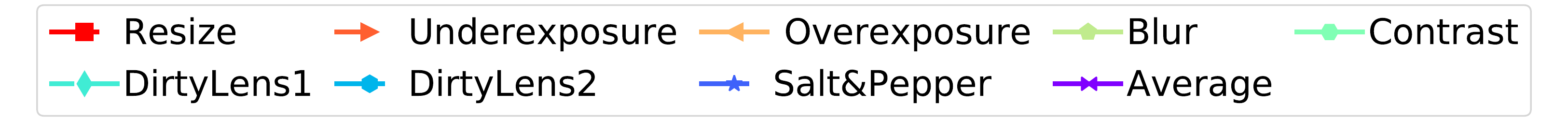}
\caption{Recognition accuracy versus acquisition conditions.}
\vspace{-9mm}
\label{fig:plots}
\end{figure}

We report the recognition performance of online platforms with respect to varying acquisition conditions in Fig.~\ref{fig:plots}. Each line represents a challenge type, except the purple line that shows the average of all challenge types. In terms of object backgrounds, white background leads to the highest recognition accuracy in both platforms as shown in Fig.~\ref{fig:plots}(a-b), which is followed by 2D textured backgrounds of kitchen and living room, and then by 3D backgrounds of office and living room. Objects are recognized more accurately in front of the white backdrop because there is not any texture or color variation in the background that can resemble other objects. The most challenging scenarios correspond to the real-world office and living room because of complex background structure. Recognition accuracy in front of 2D backdrops is higher than the real-world setups because foreground objects are more distinct when background is out of focus.

In terms of orientations, front view (0 deg) leads to the highest recognition accuracy as shown in Fig.~\ref{fig:plots}(c-d), which is expected because the objects in \texttt{CURE-OR} face forward with their most characteristic features. In contrast, these characteristic features are highly self-occluded in the overhead view, which leads to the lowest recognition performance. In case of left, right, and back views, characteristic features are not as clear as in front view but self-occlusion is not as significant as in overhead view. Therefore, these orientations lead to a medium recognition performance compared to front and overhead views. Recognition performances with respect to acquisition devices are reported in Fig.~\ref{fig:plots}(e-f), which shows that performance variation based on device types is less significant than backgrounds and orientations. However, there is still a performance difference between images obtained from different devices. Overall, Nikon D80 and Logitech C920 lead to highest recognition performance in both platforms, which highlights the importance of image quality for recognition applications.

\begin{table*}[t!]
\centering
\footnotesize
\caption{Recognition accuracy estimation performance of image feature distances in terms of Spearman correlation.}
\label{tab:accuracy_estimation}
\begin{tabular}{c|c|c|cccccccccc}
\hline
\multirow{2}{*}{\textbf{Condition}} & \multirow{2}{*}{\textbf{\begin{tabular}[c]{@{}c@{}}Feature\\Type\end{tabular}}} & \multirow{2}{*}{\textbf{Feature}}  & \multicolumn{10}{c}{\textbf{Distance Metric}} \\ \cline{4-13}  
\rule{0pt}{3.0ex}     &  & &   $\bm{l_1}$ & $\bm{l_2}$ & $\bm{{l_2}^2}$ & \textbf{SAD} & \textbf{SSAD} & \textbf{Canberra} & \textbf{Chebyshev} & \textbf{Minkowski} & \textbf{Bray-Curtis} & \textbf{Cosine} \\ \hline
\multicolumn{3}{l|}{} & \multicolumn{10}{c}{\textbf{Amazon Rekognition (AMZN)}} \\ \hline

\multirow{8}{*}{\textbf{Background}} & \multirow{5}{*}{\textbf{\begin{tabular}[c]{@{}c@{}}Hand-\\crafted\end{tabular}}} & \cellcolor{blue!10} \textbf{Color}  &0.14	&0.30	&0.29	&0.10	&0.29  &\cellcolor{blue!10}0.88	&0.01	&0.30	&0.14	&0.20  \\ 
& & \textbf{Daisy}  &0.31	&0.27	&0.26	&0.07	&0.26	&0.31	&0.40	&0.27	&0.31	&0.27 \\ 
 & & \textbf{Edge}  &0.18	&0.08	&0.12	&0.19	&0.07  &0.66	&0.04	&0.08	&0.45	&0.17 \\  
 & & \textbf{Gabor}  &0.77	&0.76	&0.76	&0.35	&0.76  &0.58	&0.71	&0.76	&0.77	&0.71  \\ 
 & & \textbf{HOG}   &0.13	&0.17	&0.16	&0.08	&0.16	&0.01	&0.12	&0.17	&0.13	&0.13\\ \cline{2-13} 
&\multirow{3}{*}{\textbf{\begin{tabular}[c]{@{}c@{}}Data-\\driven\end{tabular}}} & \textbf{VGG11}  &0.85	&0.85	&0.85	&0.10	&0.85	&0.93	&0.84	&0.85	&0.85	&0.85 \\ 
 & & \textbf{VGG13} &0.85	&0.85	&0.83	&0.01	&0.83	&0.92	&0.69	&0.85	&0.85	&0.86 \\
& &\cellcolor{blue!10} \textbf{VGG16} &0.88	&0.84	&0.84	&0.08	&0.84	&\cellcolor{blue!10} 0.94	&0.79	&0.84	&0.88	&0.85  \\ \hline

\multirow{8}{*}{\textbf{Orientation}} &\multirow{5}{*}{\textbf{\begin{tabular}[c]{@{}c@{}}Hand-\\crafted\end{tabular}}} & \textbf{Color}  &0.28	&0.41	&0.41	&0.54	&0.41  &0.04	&0.48	&0.41	&0.28	&0.16  \\  
 & & \textbf{Daisy}  & 0.45	&0.28	&0.17	&0.03	&0.17	&0.45	&0.08	&0.28	&0.45	&0.21 \\ 
 & &\cellcolor{blue!10}  \textbf{Edge}  &\cellcolor{blue!10} 0.71	&0.66	&  0.69	&0.19	&0.63 & 0.67	&0.65	&0.66	&0.65	&0.45 \\ 
 & & \textbf{Gabor}   &0.05	&0.06	&0.09	&0.39	&0.09  &0.24	&0.02	&0.06	&0.05	&0.06  \\ 
& & \textbf{HOG}  &0.19	&0.16	&0.19	&0.51	&0.19	&0.30	&0.09	&0.16	&0.19	&0.15 \\ \cline{2-13}  
&\multirow{3}{*}{\textbf{\begin{tabular}[c]{@{}c@{}}Data-\\driven\end{tabular}}} &\cellcolor{blue!10} \textbf{VGG11} &0.86	&0.92	&0.91	&0.34	&0.91	&0.69	&\cellcolor{blue!10} 0.94	&0.92	&0.86	&0.89 \\  
& & \textbf{VGG13} &0.91	&0.90	&0.84	&0.01	&0.84	&0.65	&0.78	&0.90	&0.91	&0.88 \\  
& & \textbf{VGG16} &0.88	&0.92	&0.84	&0.48	&0.84	&0.72	&0.87	&0.92	&0.88	&0.87 \\ \hline

\multicolumn{3}{l|}{} & \multicolumn{10}{c}{\textbf{Microsoft Azure (MSFT)}} \\ \hline

\multirow{8}{*}{\textbf{Background}} &\multirow{5}{*}{\textbf{\begin{tabular}[c]{@{}c@{}}Hand-\\crafted\end{tabular}}} &\cellcolor{blue!10} \textbf{Color}  &0.12	&0.13	&0.13	&0.14	&0.13 &\cellcolor{blue!10} 0.91	&0.02	&0.13	&0.12	&0.21  \\ 
&  & \textbf{Daisy}   &0.14	&0.18	&0.17	&0.01	&0.17	&0.14	&0.34	&0.18	&0.14	&0.18 \\ 
&  & \textbf{Edge}  &0.20	&0.10	&0.11	&0.27	&0.08  &0.55	&0.08	&0.10	&0.39	&0.14  \\ 
& & \textbf{Gabor}  & 0.85	&0.84	&0.84	&0.29	&0.84 &0.59	&0.80	&0.84	&0.85	&0.82  \\ 
&  & \textbf{HOG}  &0.30	&0.32	&0.31	&0.17	&0.31	&0.11	&0.18	&0.32	&0.30	&0.10 \\ \cline{2-13}  
&\multirow{3}{*}{\textbf{\begin{tabular}[c]{@{}c@{}}Data-\\driven\end{tabular}}}& \cellcolor{blue!10} \textbf{VGG11}    & 0.94	&\cellcolor{blue!10} 0.94	& 0.94	&0.13	& 0.94	&0.83	&0.90	&\cellcolor{blue!10} 0.94	& 0.94	&0.93 \\ 
& & \textbf{VGG13} &0.93	&0.92	&0.91	&0.03	&0.91	&0.86	&0.62	&0.92	&0.93	&0.93 \\
& & \textbf{VGG16} &0.91	&0.93	&0.93	&0.15	&0.93	&0.87	&0.89	&0.93	&0.91	&0.93   \\ \hline

\multirow{8}{*}{\textbf{Orientation}} &\multirow{5}{*}{\textbf{\begin{tabular}[c]{@{}c@{}}Hand-\\crafted\end{tabular}}} & \textbf{Color}   &0.28	&0.45	&0.47	&0.02	&0.47  &0.04	&0.46	&0.45	&0.28	&0.27 \\ 
 & & \textbf{Daisy}  &0.48	&0.43	&0.34	&0.24	&0.34	&0.48	&0.32	&0.43	&0.48	&0.38 \\ 
 & &\cellcolor{blue!10} \textbf{Edge}  &\cellcolor{blue!10} 0.54	&0.50	&0.51	&0.15	&0.53 &0.45	&0.47	&0.50	&0.35	&0.15 \\ 
 & & \textbf{Gabor} &0.25	&0.21	&0.18	&0.24	&0.18 &0.10	&0.23	&0.21	&0.25	&0.37  \\ 
 & & \textbf{HOG}  &0.11 &0.06	&0.11	&0.36	&0.11	&0.22	&0.13	&0.06	&0.11	&0.38 \\ \cline{2-13} 
& \multirow{3}{*}{\textbf{\begin{tabular}[c]{@{}c@{}}Data-\\driven\end{tabular}}}& \textbf{VGG11} &0.38	&0.46	&0.50	&0.03	&0.50	&0.34	&0.42	&0.46	&0.38	&0.43 \\ 
& & \textbf{VGG13} &0.52	&0.48	&0.47	&0.15	&0.47	&0.44	&0.43	&0.48	&0.52	&0.51 \\
& &\cellcolor{blue!10} \textbf{VGG16} &0.43	&0.46	&0.48	&\cellcolor{blue!10} 0.71	&0.48	&0.46	& 0.53	&0.46	&0.43	&0.44   \\ \hline
 \end{tabular}
\end{table*}

\begin{figure*}[htbp!]
\begin{minipage}[b]{0.20\linewidth}
  \centering
\includegraphics[width=\textwidth]{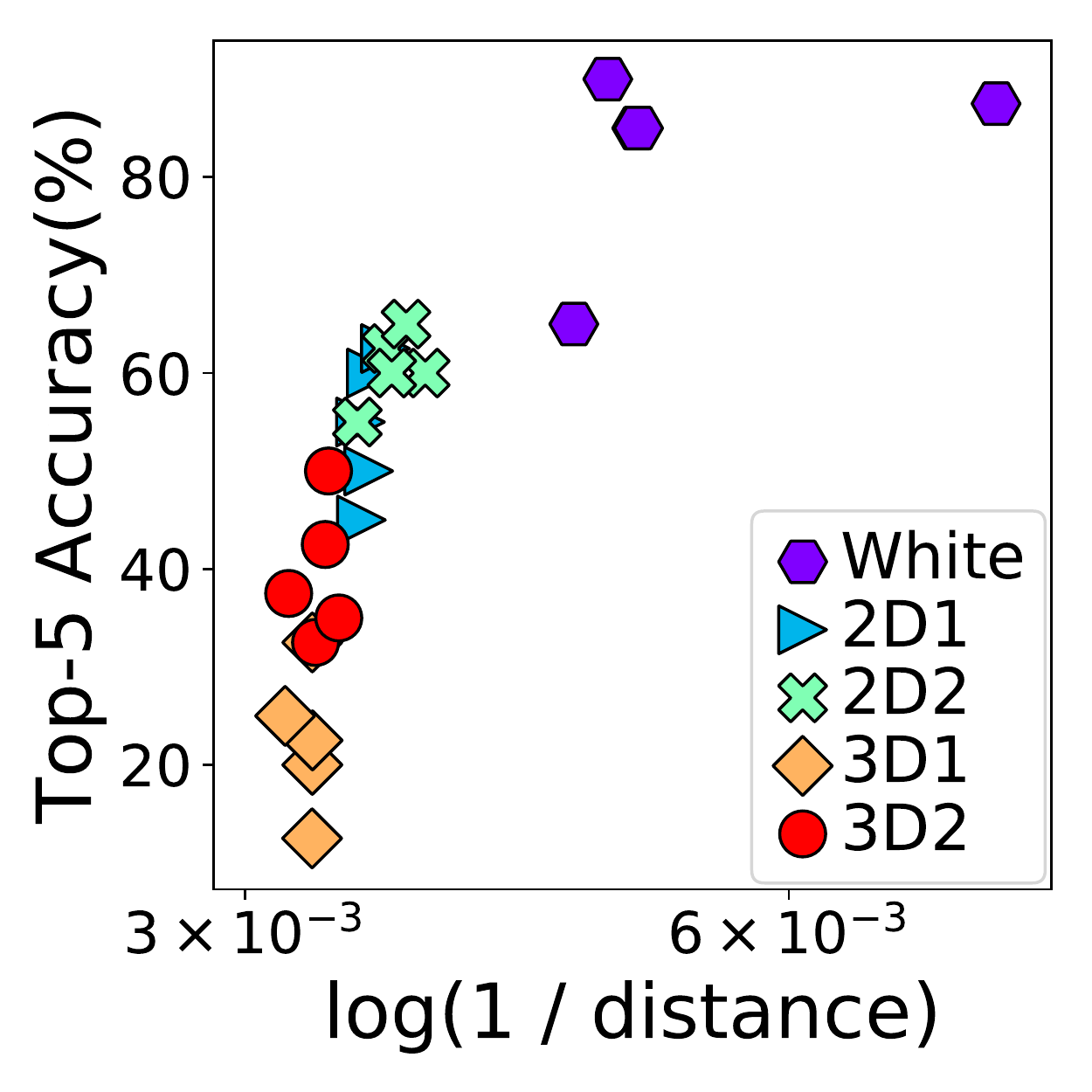}
  \centerline{\footnotesize{(a)
  {\tabular[t]{@{}l@{}} AMZN Background\\- VGG16 Canberra \endtabular}}}
\end{minipage}
\hfill
\begin{minipage}[b]{0.20\linewidth}
  \centering
\includegraphics[width=\textwidth]{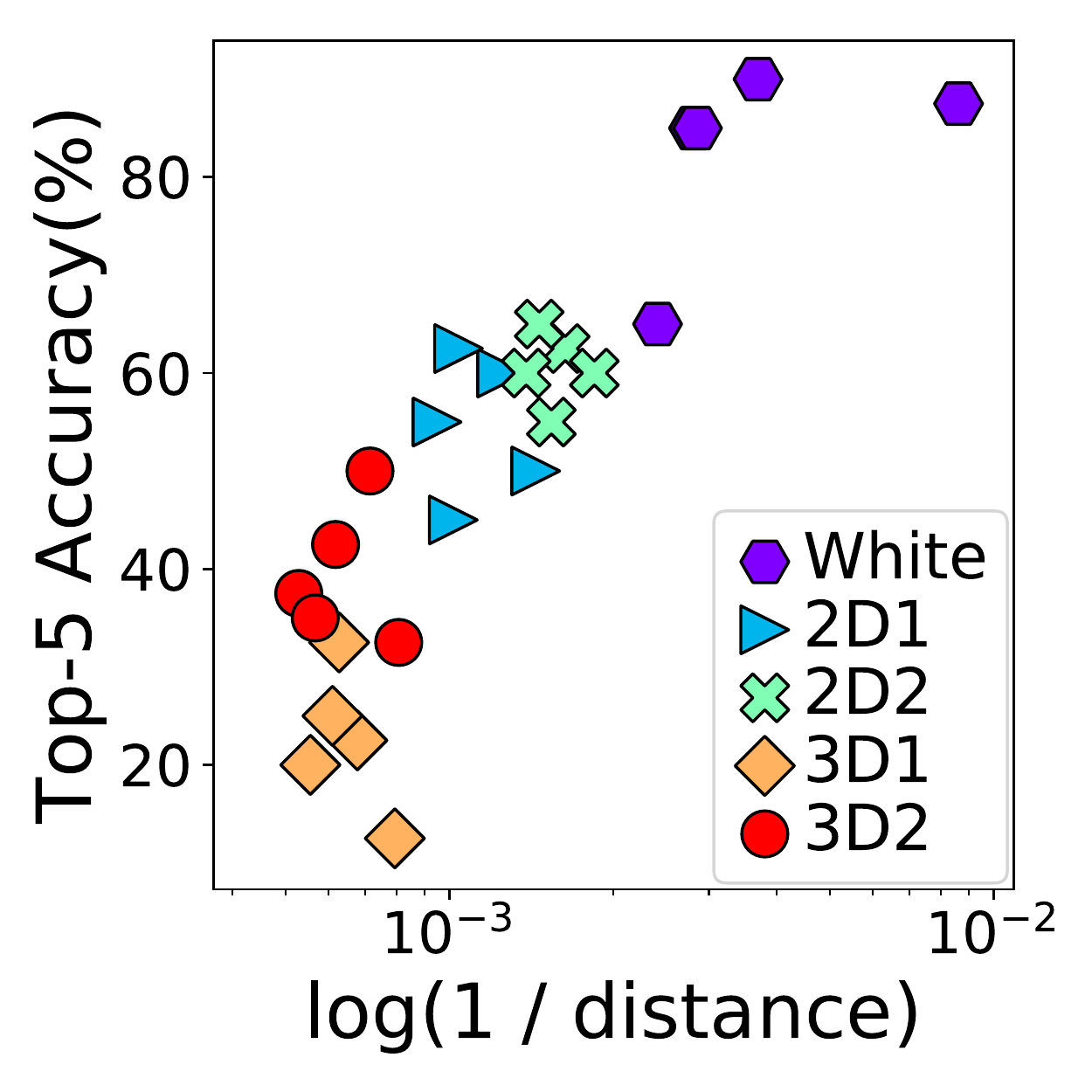}
  \centerline{\footnotesize{(b) 
  {\tabular[t]{@{}l@{}} AMZN Background\\- Color Canberra \endtabular}}}
\end{minipage}
\hfill
\begin{minipage}[b]{0.20\linewidth}
  \centering
\includegraphics[width=\textwidth]{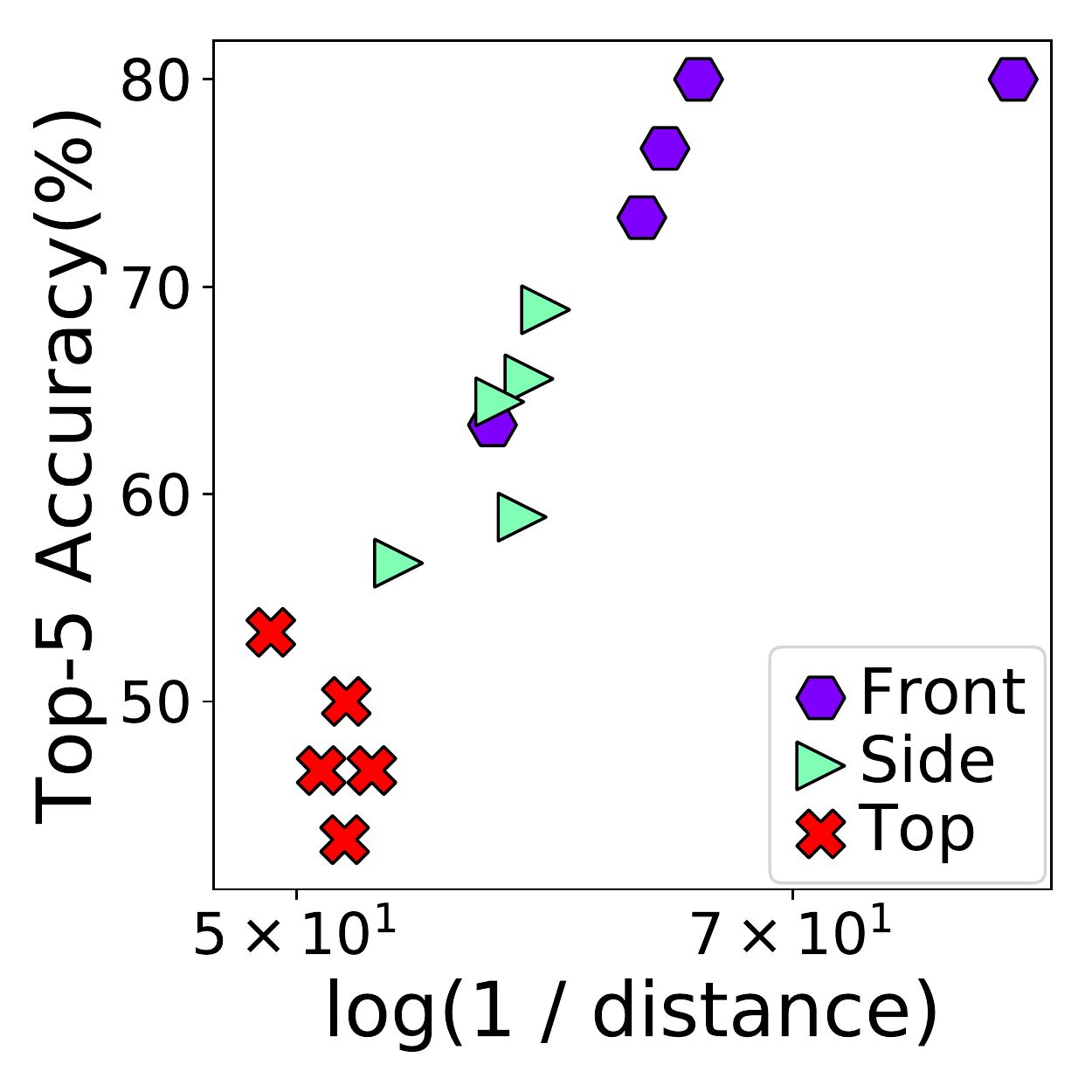}
  \centerline{\footnotesize{(c) 
  {\tabular[t]{@{}l@{}} AMZN Orientation\\- VGG11 Chebyshev \endtabular}}}
\end{minipage}
\hfill
\begin{minipage}[b]{0.20\linewidth}
  \centering
\includegraphics[width=\textwidth]{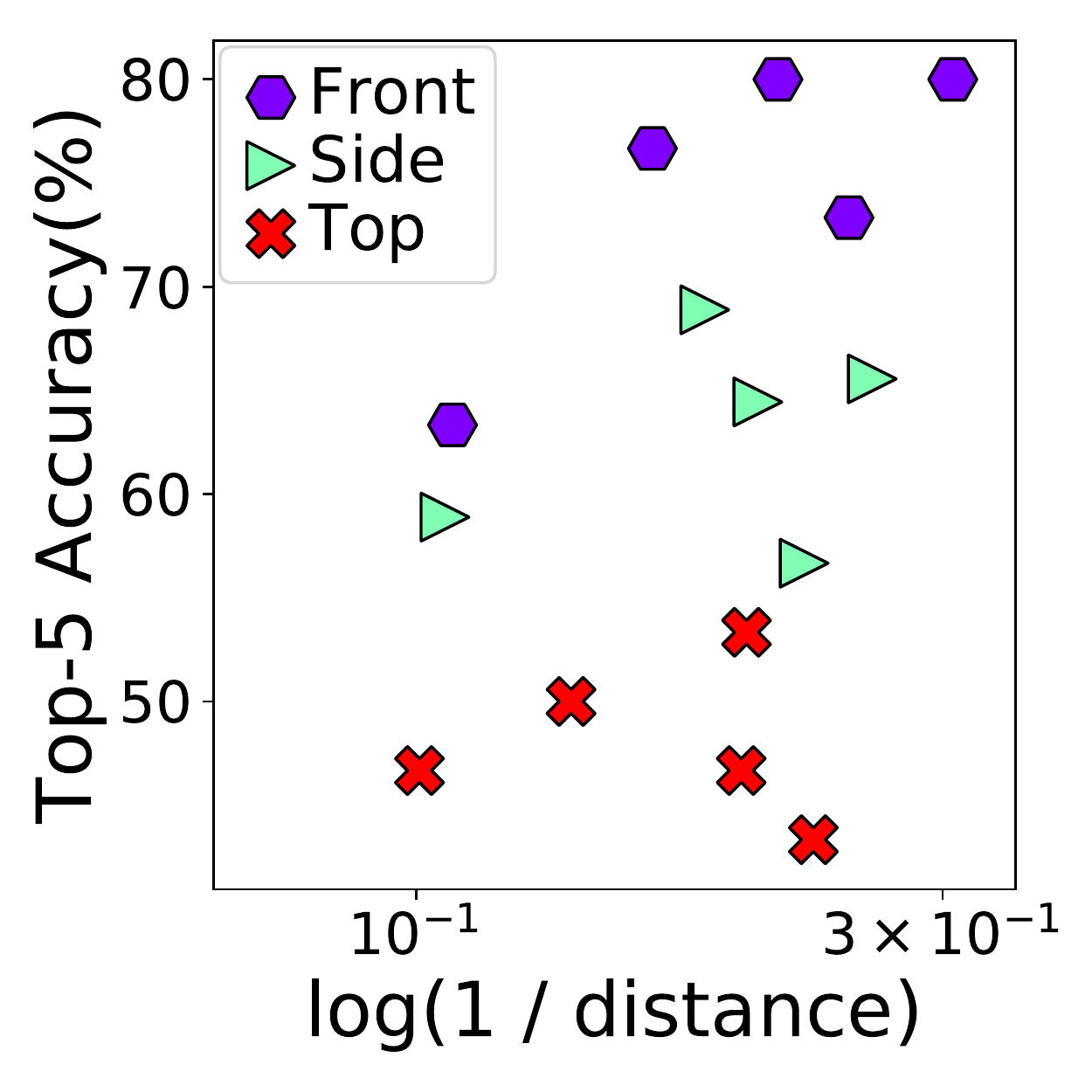}
  \centerline{\footnotesize{(d) 
  {\tabular[t]{@{}l@{}} AMZN Orientation\\- Edge l1 \endtabular}}}
\end{minipage}
\begin{minipage}[b]{0.20\linewidth}
  \centering
\includegraphics[width=\textwidth]{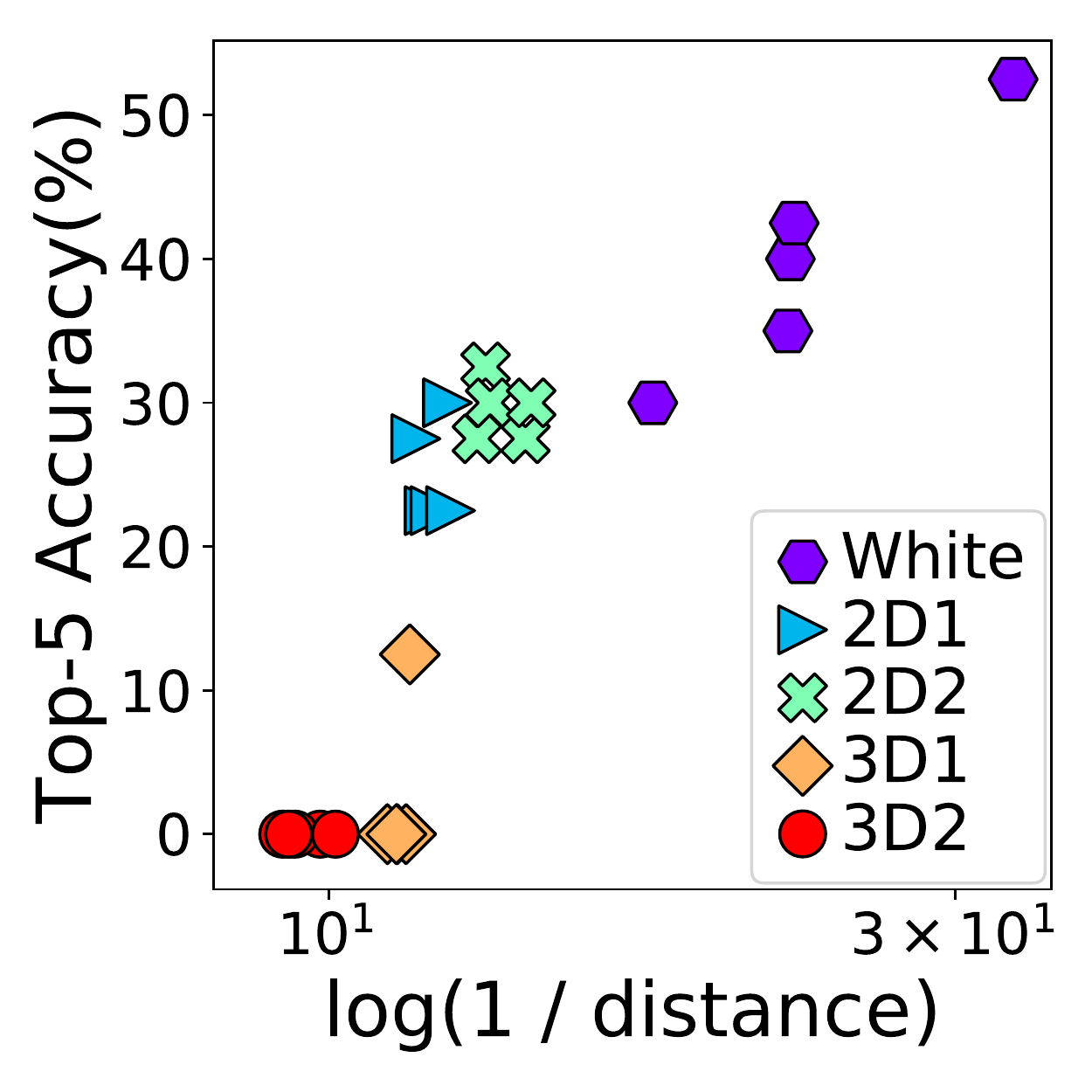}
  \centerline{\footnotesize{(e) 
  {\tabular[t]{@{}l@{}} MSFT Background\\- VGG11 Minkowski \endtabular}}}
\end{minipage}
\hfill
\begin{minipage}[b]{0.20\linewidth}
  \centering
\includegraphics[width=\textwidth]{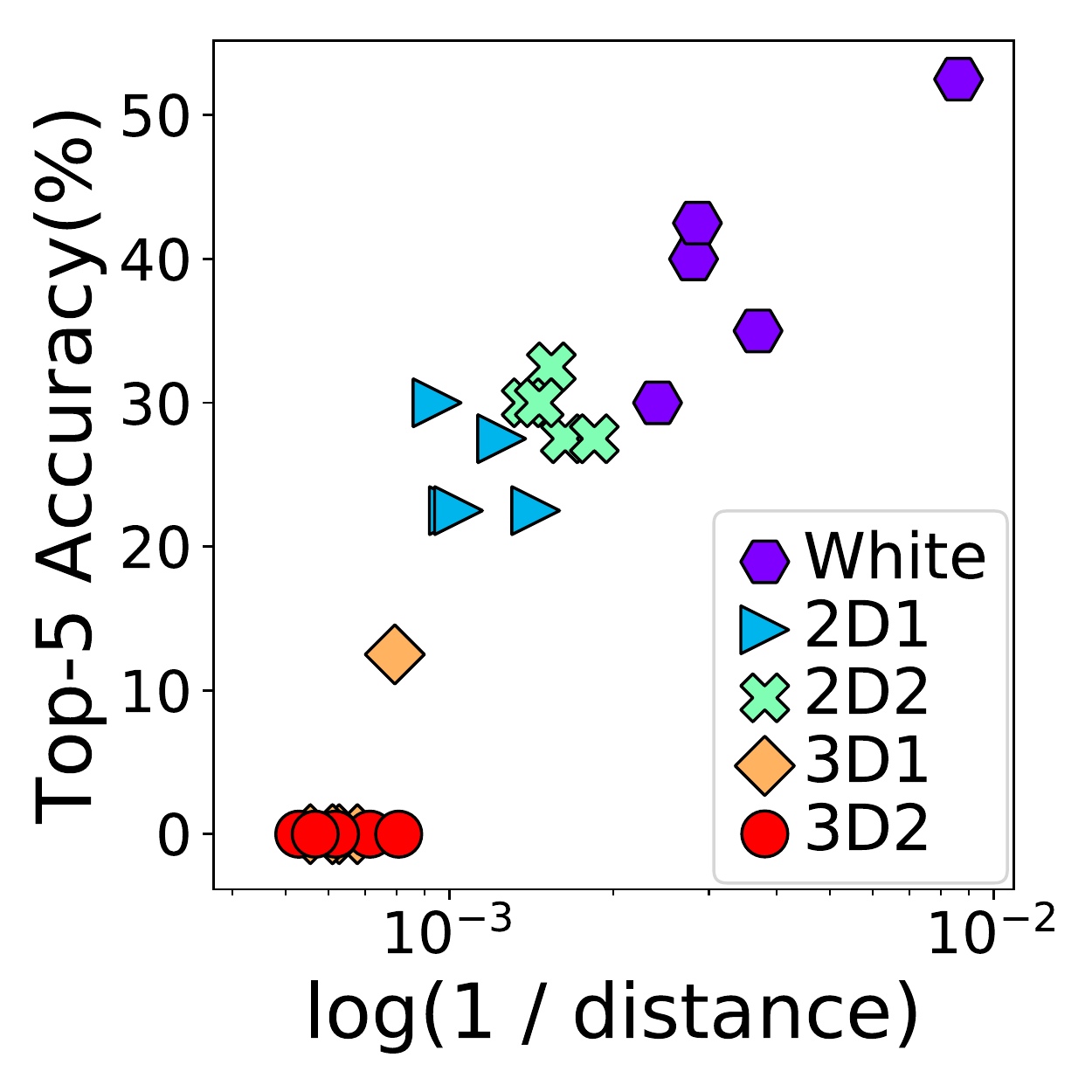}
  \centerline{\footnotesize{(f) 
  {\tabular[t]{@{}l@{}} MSFT Background\\- Color Canberra \endtabular}}}
\end{minipage}
\hfill
\begin{minipage}[b]{0.20\linewidth}
  \centering
\includegraphics[width=\textwidth]{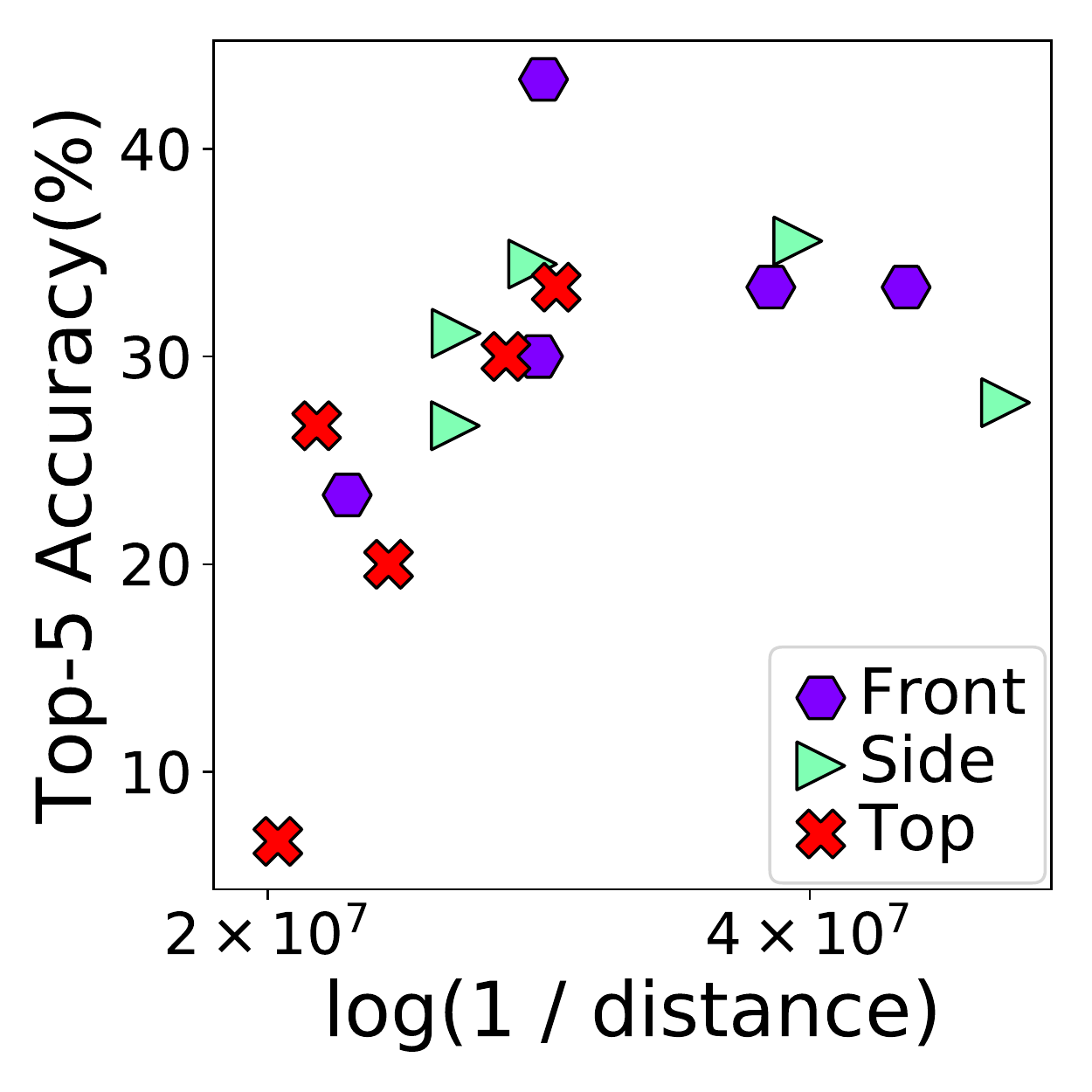}
  \centerline{\footnotesize{(g) 
  {\tabular[t]{@{}l@{}} MSFT Orientation\\- VGG16 SAD \endtabular}}}
\end{minipage}
\hfill
\begin{minipage}[b]{0.20\linewidth}
  \centering
\includegraphics[width=\textwidth]{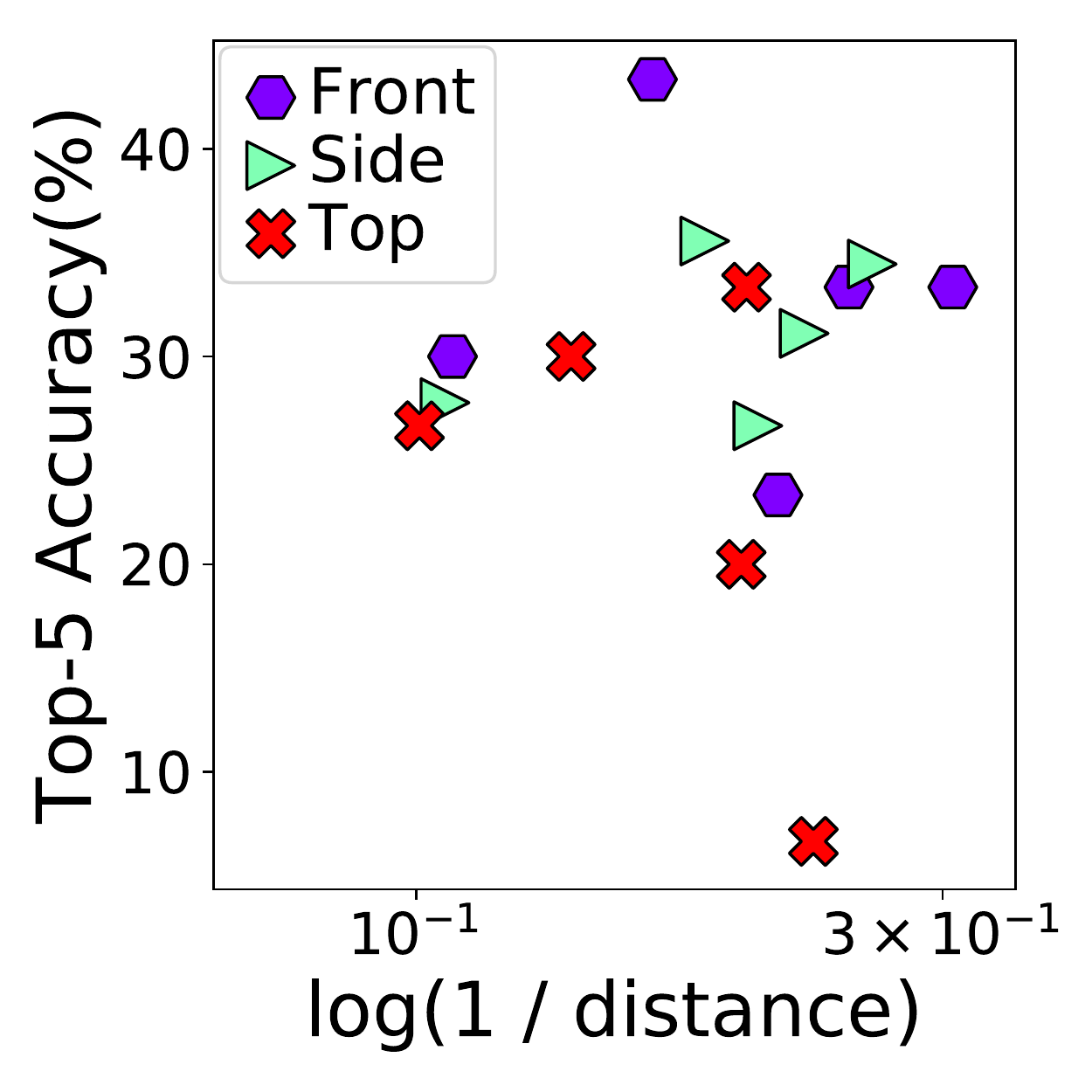}
  \centerline{\footnotesize{(h) 
  {\tabular[t]{@{}l@{}} MSFT Orientation\\- Edge l1 \endtabular}}}
\end{minipage}
\centering
\vspace{-2 mm}
\caption{Scatter plots of top hand-crafted and data-driven recognition accuracy estimation methods.}
\vspace{-5mm}
\label{fig:accuracy_scatter}
\end{figure*}

\vspace{-2mm}
\section{Recognition Performance Estimation under Multifarious Conditions}
\vspace{-2mm}
\label{sec:main_estimation}
Based on the experiments reported in Section \ref{sec:main_recognition}, the reference configuration that leads to the highest recognition performance is front view, white background, and Nikon DSLR. We conducted two experiments to estimate the recognition performance with respect to changes in background and orientation. We utilized the 10 common objects of both platforms for direct comparison. In the \emph{background experiment}, we grouped images captured with a particular device (5) in front of a specific background (5), which leads to 25 image groups with front and side views of the objects. In the \emph{orientation experiment}, we grouped images captured with a particular device (5) from an orientation (3) among front, top, and side views, which leads to 15 image groups with images of the objects in front of white, living room, and kitchen backdrops. For each image group, we obtained an average recognition performance per recognition platform and an average feature distance between the images in the group and their reference image. Finally, we analyzed the relationship between recognition accuracy and feature distance with correlations and scatter plots. We extracted commonly used handcrafted and data-driven features as follows: \vspace{-2mm}
\begin{itemize}[label=\textcolor{orange}{\FilledSmallSquare},leftmargin=*]
  \setlength\itemsep{-0.4 em}
    \item  \texttt{Color:} Histograms of color channels in RGB.
    \item  \texttt{Daisy:} Local image descriptor based on convolutions of gradients in specific directions with Gaussian filters \cite{Tola2010}.
    \item  \texttt{Edge:} Histogram of vertical, horizontal, diagonal, and non-directional edges. 
    \item  \texttt{Gabor:} Frequency and orientation information of images extracted through Gabor filters.
    \item  \texttt{HOG:} Histogram of oriented gradients over local regions.
    \item  \texttt{VGG:} Features obtained from convolutional neural networks that are based on stacked $3 \times 3$ convolutional layers \cite{Simonyan2014}. The VGG index indicates the number of weighted layers in which last three layers are fully connected layers.
    \end{itemize}
We calculated the distance between features with  $l_1$ norm, $l_2$ norm, ${l_2}^2$ norm, sum of absolute differences (SAD), sum of squared absolute differences (SSAD), weighted $l_1$  norm (Canberra), $l_\infty$ norm (Chebyshev), Minkowski distance, Bray-Curtis dissimilarity, and Cosine distance. We report the recognition accuracy estimation performance in Table~\ref{tab:accuracy_estimation} in terms of Spearman correlation between top-5 recognition accuracy scores and feature distances. We highlight the top data-driven and hand-crafted methods with light blue for each recognition platform and experiment. 

In the \emph{background experiment}, color characteristics of different backgrounds are distinct from each other as observed in Fig.~\ref{fig:images}. In terms of low level characteristic features including \texttt{Daisy}, \texttt{Edge}, and \texttt{HOG}, edges in the backgrounds can distinguish highly textured backgrounds from less textured backgrounds. However, edges would not be sufficient to distinguish lowly textured backgrounds from each other. Moreover, edges of the foreground objects can dominate the feature representations and mask the effect of changes in the backgrounds. To distinguish differences in backgrounds overlooked by edge characteristics, frequency and orientation characteristics can be considered with \texttt{Gabor} features. Data-driven methods including \texttt{VGG} utilize all three channels of images while extracting features, which can give them an inherent advantage with respect to solely color or structure based methods. Overall, data-driven method \texttt{VGG} leads to the highest performance in the \emph{background experiment} for both recognition platforms. In terms of hand-crafted features, \texttt{color} leads to the highest performance followed by \texttt{Gabor} whereas edge-based methods result in inferior performance.

Distinguishing changes in orientation is more challenging compared to backgrounds because region of interest is limited to a smaller area. Therefore, overall recognition accuracy estimation performances are lower for orientations compared to backgrounds as reported in Table~\ref{tab:accuracy_estimation}. Similar to the \emph{background experiment}, \texttt{VGG} architectures lead to the highest performance estimation in the \emph{orientation experiment}. However, hand-crafted methods are dominated by \texttt{edge} features instead of \texttt{Gabor} representations. We show the scatter plots of top performing data-driven and hand-crafted methods in Fig.~\ref{fig:accuracy_scatter} in which x-axis corresponds to average distance between image features and y-axis corresponds to top-5  accuracy. Image groups corresponding to different configurations are more distinctly clustered in terms of background as observed in Fig.~\ref{fig:accuracy_scatter}(a-b, e-f). In terms of orientation, \texttt{VGG} leads to a clear distinction of configurations for Amazon Rekognition as observed in Fig.~\ref{fig:accuracy_scatter}(c) whereas image groups are overlapping in other experiments as shown in Fig.~\ref{fig:accuracy_scatter}(d, g-h). Clustering configurations is more challenging in the \emph{orientation} experiment because it is not even possible to easily separate orientation configurations based on their recognition accuracy.

\vspace{-2mm}
\section{Conclusion}
\vspace{-2mm}
\label{sec:conc}
In this paper, we analyzed the robustness of recognition platforms and reported that object background can affect recognition performance as much as orientation whereas tested device types have minor influence on recognition. We also introduced a framework to estimate recognition
performance variation and showed that color-based features capture background variations, edge-based features capture orientation variations, and data-driven features capture both background and orientation variations in a controlled setting. Overall, recognition performance can significantly change depending on the acquisition conditions, which highlights the need for more robust platforms that we can confide in our daily lives. Estimating recognition performance with feature similarity-based metrics can be helpful to test the robustness of algorithms before deployment. However, the applicability of such estimation frameworks can drastically increase if we design no-reference approaches that can provide a recognition performance estimation without a reference image similar to the no-reference algorithms in image quality assessment field.

\newpage







\end{document}